\documentclass{article}

\usepackage[final]{nips_2016}

\usepackage[utf8]{inputenc} 
\usepackage[T1]{fontenc}    
\usepackage{hyperref}       
\usepackage{url}            
\usepackage{booktabs}       
\usepackage{amsfonts}       
\usepackage{nicefrac}       
\usepackage{microtype}      
\usepackage{graphicx}
\title{Long Timescale Credit Assignment in Neural Networks with External Memory}

\author{
  Steven S. Hansen \\
  Department of Psychology\\
  Stanford University\\
  Stanford, CA 94303 \\
  \texttt{sshansen@stanford.edu} \\
}

\begin{document}

\maketitle

\section{Introduction}
Neural networks with external memories (NNEM),such as the Neural Turing Machine [1] and Memory Networks [2], have often been compared to the human hippocampus in their ability to maintain episodic information over long timescales [3]. But so far, these networks have only been trained on tasks requiring memory storage comparable to a few minutes, whereas the hippocampus stores information on the order of years. It is well known that this sort of long-term episodic storage is vital for overcoming the problem of catastrophic interference [4], which is central to the challenge of continual learning.

While some of this discrepancy can be attributed to the short-term nature of current benchmark tasks, a major bottleneck comes form the reliance on using back-propagation-through-time to learn the embedding function that maps the high-dimensional observations onto lower dimensional representations suitable for storage in the external memory. This is problematic because memories might be used an extremely long time after the embedding function was called, and back-propagation-through-time would require us to hold onto the forward activations of this function for the entire duration if we want to back-propagate through it. 

Considering that in a continual learning setup, an external memory store on the order of hundreds of thousands, or even millions, of separate embeddings could be needed, storing all of the intermediate embedding activations is woefully intractable. A naive solution would be to simply store all of the memories in observation space in addition to embedding space and recompute the embeddings as needed for credit assignment, but a significant computational burden would result from these redundant forward-passes. More importantly, the high dimensionality of the observation space would make storing these additional memories take an unreasonable amount of memory. Indeed, Blundell et al [5] noted that even in the relatively modest Atari domain, storing a few million observations in pixel-space would require upwards of 300 gigabytes\footnote{While their work utilizes a long timescale external memory, it does not address the problems raised here, as the embedding function was not optimized for usage in an external memory. Rather, it was either a random projection or pre-trained on a separate objective. Indeed, the low performance of the latter suggests that end-to-end optimization might be necessary for adaptive embedding functions to be worthwhile in this context.}.

\section{Methods}
\subsection{Long Timescale Credit Assignment in NNEMs}

Credit assignment in traditional recurrent neural networks usually involves back-propagating through a long chain of tied weight matrices. The length of this chain scales linearly with the number of time-steps as the same network is run at each time-step. This creates many problems, such as vanishing gradients, that have been well studied [7]. In contrast, a NNEM's architecture recurrent activity doesn't involve a long chain of activity (though some architectures such as the NTM do utilize a traditional recurrent architecture as a controller). Rather, the externally stored embedding vectors are used at each time-step, but no messages are passed from previous time-steps. This means that vanishing gradients aren't a problem, as all of the necessary gradient paths are short. However, these paths are extremely numerous (one per embedding vector in memory) and reused for a very long time (until it leaves the memory). Thus, the forward-pass information of each memory must be stored for the entire duration of the memory. This is problematic as this additional storage far surpasses that of the actual memories, to the extent that large memories on infeasible to back-propagate through in high dimensional settings.

\begin{figure}[h]
  \centering
  \fbox{\includegraphics[scale=.32]{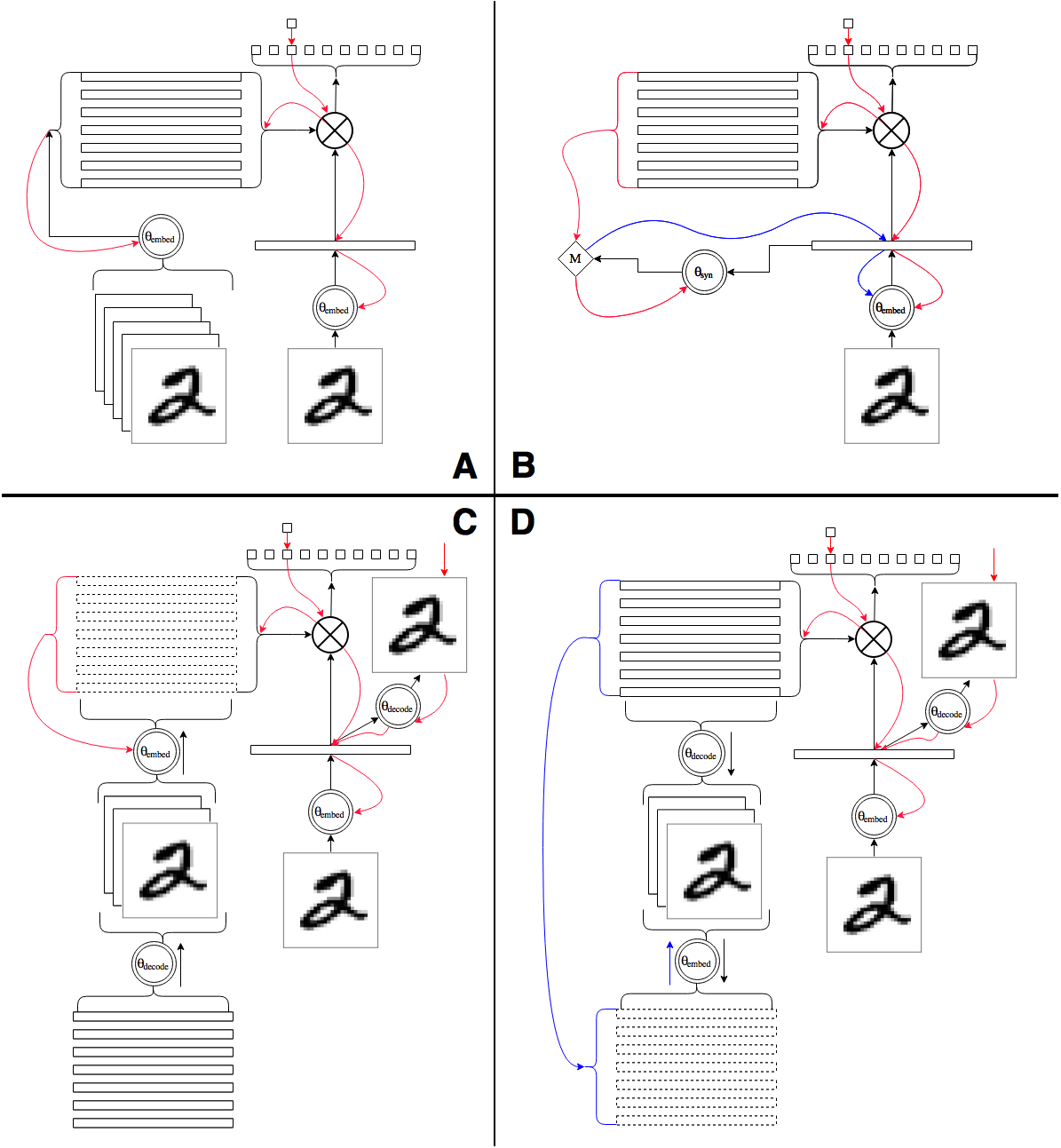}}
  \caption{The computational graphes of a NNEM with differing credit assignment mechanisms. Forward-pass information is shown in black, true gradients in red, and approximate gradients in blue. A) The standard method for back-propagating through an external memory. Note that the left half of this graph must be frozen until the memory leaves. B) Synthetic gradient model, M, assigns credit upon first entering the memory, and then itself learning from the older memories' gradient. C) Exact gradient reinstatement. The dashed memories are online recalculations. D) Approximate gradient reinstatement. The memories used for inference and credit assignment are different.}
  \label{fig:main}
\end{figure}

\subsection{Synthetic Gradients}

Synthetic gradients are a recent technique created to deal with situation whether a part of a network is `blocked' from further usage [6], such as in the case previously discussed whereby information must be stored until all gradient calculations are complete. Their way around this problem is to treat gradient signal as just another function to be approximated. Namely, a function mapping from an activation vector, and any contextual information, onto to the error gradient w.r.t. that activation vector. This auxiliary network can be used to calculate an estimate of the gradient before the true gradient is available. This synthetic gradient producing network can be trained in a purely supervised manner by comparing its estimates to the true gradient. 

As shown in Figure \ref{fig:main}B, we can apply this approach to NNEMs by updating the embedding network with a synthetic gradient immediately after the embedding vector is calculated. True gradient signals are available for all of the previously embedded memories, and these can be used to train the synthetic gradient network.

\subsection{Reinstatement for Credit Assignment}

One way to get around the need to hold onto forward-pass information is to recalculate the forward-pass whenever gradient information is available. However, as previously mentioned, even the observations are too large to store in our domain of interest, which prevents a direct reinstatement of a forward pass from occurring. Instead, we rely on a learned autoencoder to reinstate the observation, and then use the embedding network to recalculate the forward-pass. Since the recalculated embedding vector is unlikely to perfectly match the one stored in memory, it is non-obvious how to utilize error gradient w.r.t. the vector in memory.

There are two distinct possibilities: ignore the mismatch and directly use the memory's gradient, or use the fresh embedding instead of the memory during inference. These two methods are illustrated in Figure \ref{fig:main}C and D, and while they aren't mutually exclusive, they are treated as such for the purposes of this paper, to aid in interpreting the initial results. The latter option is a true gradient signal whereas the former is only an approximation, but it relies on having a good enough reconstruction to not harm the inference process.

\section{Results}

To test our model, we used the standard MNIST benchmark, the objective of which is to correctly classify images of numerals by the integer they represent. While this task is trivial for modern networks with the appropriate architecture, we use it here as it represents one of the simplest tasks that could, under certain conditions, still encounter the credit assignment problem previously described.

\begin{figure}[h]
  \centering
  \fbox{\includegraphics[scale=.5]{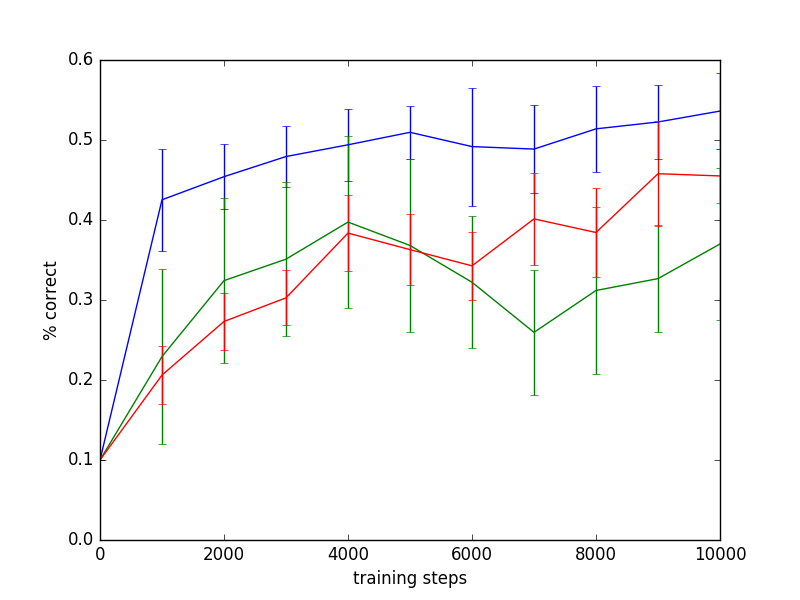}}
  \caption{Performance results averaged across 10 runs. Exact gradient reinstatement is shown in blue, approximate gradient reinstatement is shown in green, and synthetic gradients are shown in red. All accuracy results are from a seperate validation set.}
  \label{fig:res}
\end{figure}

Since our aim is to tractably train embedding functions within a network with an external memory, we constructed a simple NNEM consisting of an embedding network and an external store that holds the embeddings of the 5000 most recently encountered images along with their correct labels. Classifications are made by taking the cosine similarity between the embedding of the current image and all of the embeddings in memory, and using the normalized result to weight class labels associated with the memorized embeddings. As our focus is on weight updates derived from the memorized embeddings, we stop the gradient going into the embedding function coming from the current image, as these are sufficient for good performance in this simplified setting. In addition, we found that completely unsupervised learning of an autoencoder was sufficient to yield decent performance. Since we're primarily interested in getting the supervised signal to travel through the stored embeddings, we only allow the reconstruction error to modify the decoder's weights, as the decoder isn't used for inference like the encoder/embedding function is.

The results are shown in Figure\ref{fig:res}. All conditions utilize the same network architecture\footnote{a single hidden layer for each piece of the network, with dimension 256. ADAM was used for optimization with a learning rate of 1e-4 elsewhere}  and hyper-parameters, different only in their mechanism for credit assignment. The reinstatement method with exact gradients performed the best, but requires the most additional computation. Synthetic gradient performance is perhaps lower than expected, but it is possible that this is due to the network architecture. The surrounding memories affect the gradient of any individual memory, but these can't be an input to the synethic gradient network, as future memories aren't observable. The approximate gradient version of the reinstatement method was the most prone to divergence, which makes sense given the approximation involved.

\section{Future Directions}

The absolute performance on this MNIST task isn't useful in itself, as the performance was purposely sabotaged to isolate the novel mechanisms under discussion. However, while the relative performance changes demonstrate the promise of reinstatement based credit assignment, the contrived domain necessitates further work to confirm these results on large scale problems. One possibility would be tackle reinforcement learning problems like Atari, using a variant of the model-free episodic control network [5] modified such that the embedding function is learned on-line. While the exact modifications needed to make that setting compatible with this approach to credit assignment is outside the scope of this paper, it is clear that this domain is one of the few present in the current literature that calls for a long term episodic store.

In a more complex environment, additional nuances emerge which might improve credit assignment performance. Given the reliance on accurate decoding for proper reinstatement, pre-training the autoencoder offline might be crucial. Vanilla autoencoders also tend to overfit, which might slow performance when used to assign credit to new examples. Variational autoencoders have been shown to be more robust [8] and also provide uncertainty estimates that could be useful when deciding which embeddings to discard. A hybrid approach utilizing synthetic gradients and both forms of AE based credit assignment might also be worthwhile, as these approaches appear to have complementary weaknesses.

While the focus has been on credit assignment, adaptive long-term episodic memories raise other concerns, such as the fact that stored embedding go `stale' rapidly, as the embedding function's parameters change. If the rate of change is slow enough, then operating the memories as a queue (first in, first out) should limit the impact this has on performance. But in many setups, embeddings are discarded as a function of their usage [5] such that frequently needed embeddings are kept around long enough for staleness to become problematic. While only speculative, an autoencoder could also be used to `freshen' these embeddings by re-encoding them using the current parameters. 

\section*{References}
\small

[1] Graves, A., Wayne, G., \& Danihelka, I. (2014). Neural turing machines. {\it arXiv preprint arXiv}:1410.5401.

[2] Weston, J., Chopra, S., \& Bordes, A. (2014). Memory networks. {\it arXiv preprint arXiv}:1410.3916.

[3] Graves, A., Wayne, G., Reynolds, M., Harley, T., Danihelka, I., Grabska-Barwińska, A., ... \& Badia, A. P. (2016). Hybrid computing using a neural network with dynamic external memory. {\it Nature}.

[4] McClelland, J. L., McNaughton, B. L., \& O'Reilly, R. C. (1995). Why there are complementary learning systems in the hippocampus and neocortex: insights from the successes and failures of connectionist models of learning and memory. {\it Psychological review}, 102(3), 419.

[5] Blundell, Charles, et al. "Model-free episodic control." {\it arXiv preprint arXiv:1606.04460} (2016).

[6] Jaderberg, M., Czarnecki, W. M., Osindero, S., Vinyals, O., Graves, A., \& Kavukcuoglu, K. (2016). Decoupled neural interfaces using synthetic gradients. {\it arXiv preprint arXiv}:1608.05343.

[7]Hochreiter, S. (1998). The vanishing gradient problem during learning recurrent neural nets and problem solutions. {\it International Journal of Uncertainty, Fuzziness and Knowledge-Based Systems}, 6(02), 107-116.

[8]Kingma, D. P., \& Welling, M. (2013). Auto-encoding variational bayes. {\it arXiv preprint arXiv}:1312.6114.
\end{document}